\documentclass{article}
\usepackage{spconf,amsmath,graphicx}
\usepackage{enumitem}
\setlist{nosep, leftmargin=14pt}
\usepackage{mwe} 
\usepackage{amsmath}
\usepackage{amssymb}
\usepackage{algorithmic}
\usepackage{array}
\usepackage{url}
\usepackage{multirow}
\usepackage{hyperref}
\usepackage{subfig}
\newcommand{\orcid}[1]{\href{https://orcid.org/#1}{\includegraphics[width=10pt]{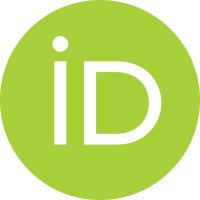}}}

\title{AWGUNet: Attention-aided Wavelet Guided U-net for nuclei segmentation in histopathology images}
\name{Ayush Roy$^{1,\dagger}$\orcid{0000-0002-9330-6839}\thanks{$^{\dagger}$ These authors contributed equally to this work.} \qquad
     Payel Pramanik$^{2,\dagger}$\orcid{0000-0002-6086-0681} \qquad
     Dmitrii Kaplun$^{3,4,\star}$\orcid{0000-0003-2765-4509}\thanks{$^{\star}$ Corresponding author.} \quad
     Sergei Antonov $^{3,4}$ \orcid{0000-0002-3370-7032}\qquad
     Ram Sarkar$^{2}$\orcid{0000-0001-8813-4086}}
\address{$^1$ Department of Electrical Engineering, Jadavpur University, Kolkata, India. \\
        $^2$ Department of Computer Science and Engineering, Jadavpur University, Kolkata, India. \\
        $^3$ Artificial Intelligence Research Institute, China University of Mining and Technology,  Xuzhou, China.\\
        $^4$ Department of Automation and Control Processes, Saint Petersburg Electrotechnical University "LETI",\\
            Saint Petersburg, Russia.    
        }
\begin{document}
%
\maketitle
\begin{abstract}
Accurate nuclei segmentation in histopathological images is crucial for cancer diagnosis. Automating this process offers valuable support to clinical experts, as manual annotation is time-consuming and prone to human errors. However, automating nuclei segmentation presents challenges due to uncertain cell boundaries, intricate staining, and diverse structures. In this paper, we present a segmentation approach that combines the U-Net architecture with a DenseNet-121 backbone, harnessing the strengths of both to capture comprehensive contextual and spatial information. Our model introduces the Wavelet-guided channel attention module to enhance cell boundary delineation, along with a learnable weighted global attention module for channel-specific attention. The decoder module, composed of an upsample block and convolution block, further refines segmentation in handling staining patterns. The experimental results conducted on two publicly accessible histopathology datasets, namely Monuseg and TNBC, underscore the superiority of our proposed model, demonstrating its potential to advance histopathological image analysis and cancer diagnosis. The code is made available at: \url{https://github.com/AyushRoy2001/AWGUNET}
\end{abstract}
\section{Introduction}
\label{sec:intro}
Nuclei and cell micro-environments provide vital diagnostic information, particularly as cancer transforms normal cells, altering characteristics such as nuclei count, size, and morphology. Accurate segmentation of nuclei in stained pathological images is crucial for pathological research, which is typically a time-consuming task for radiologists. Automating this segmentation becomes valuable given the high density of entire slide images. However, the diversity in uncertain cell boundaries continues to challenge automated nuclei segmentation in clinical pathology analysis. While machine learning techniques have shown progress in medical image segmentation, such methods struggle to extract highly representative features and suffer from limited neighborhood receptive fields. In recent years, deep learning-based methods~\cite{naylor2018segmentation,kanadath2023multilevel,das2023deep} have advanced rapidly, finding widespread application in medical image segmentation. Convolutional Neural Networks (CNNs) based models, in particular, have demonstrated exceptional performance in this domain. However, due to the presence of uncertain cell boundaries, complex staining, and intricate and heterogeneous histopathological structure, it becomes challenging to segment nuclei cells accurately. Accurate localization of nuclei cell boundaries and handling intricate staining patterns remain fundamental challenges for this task. Additionally, ensuring the model's robustness across a wide spectrum of cell morphologies and image variations presents another significant challenge. \par

\textbf{Motivation.} With the emergence of the U-Net\cite{ronneberger2015u} model, the encoder-decoder structure has become the go-to choice for image segmentation, including medical images. U-Net utilizes skip connections to effectively integrate context information during network training, achieving deep feature fusion across multiple scales. To enhance U-Nets' ability, various approaches have emerged, such as the 3D U-Net for 3D image segmentation, the residual-based ResUnet++, etc. Attention mechanisms have been a key focus in enhancing the performance of U-Net for medical image segmentation, leading to the development of models like the attention U-Net\cite{oktay2018attention}. These methods leverage attention mechanisms to extract discriminative features, allowing the model to concentrate on the region of interest and improve image segmentation. While most existing medical image segmentation networks based on U-Net aim to enhance contextual information for feature extraction, the importance of spatial features, particularly edge information, should not be overlooked. \par

\textbf{Contributions.} To capture both the contextual and spatial information, we propose a model that adopts a U-Net architecture with a DenseNet-121 backbone in this work. This combination leverages the strengths of both structures, offering a powerful foundation for accurate segmentation in histopathological images. To capture the edge information for spatial guidance, we introduce a Wavelet Guided Channel Attention Module (WGCAM), which improves the model's ability to delineate cell boundaries with precision. We further enhance the model performance with a learnable weighted Global Average Pooling (lw-GAP), which provides channel-specific attention. Additionally, the decoder module, featuring the upsample block and convolution block, refines segmentation, especially regarding staining patterns and extremely small regions of interest. 

\section{Methodology}
\label{sec:method}
\begin{figure}[htb]
    \centering
    \includegraphics[width=0.95\linewidth, height=6cm]{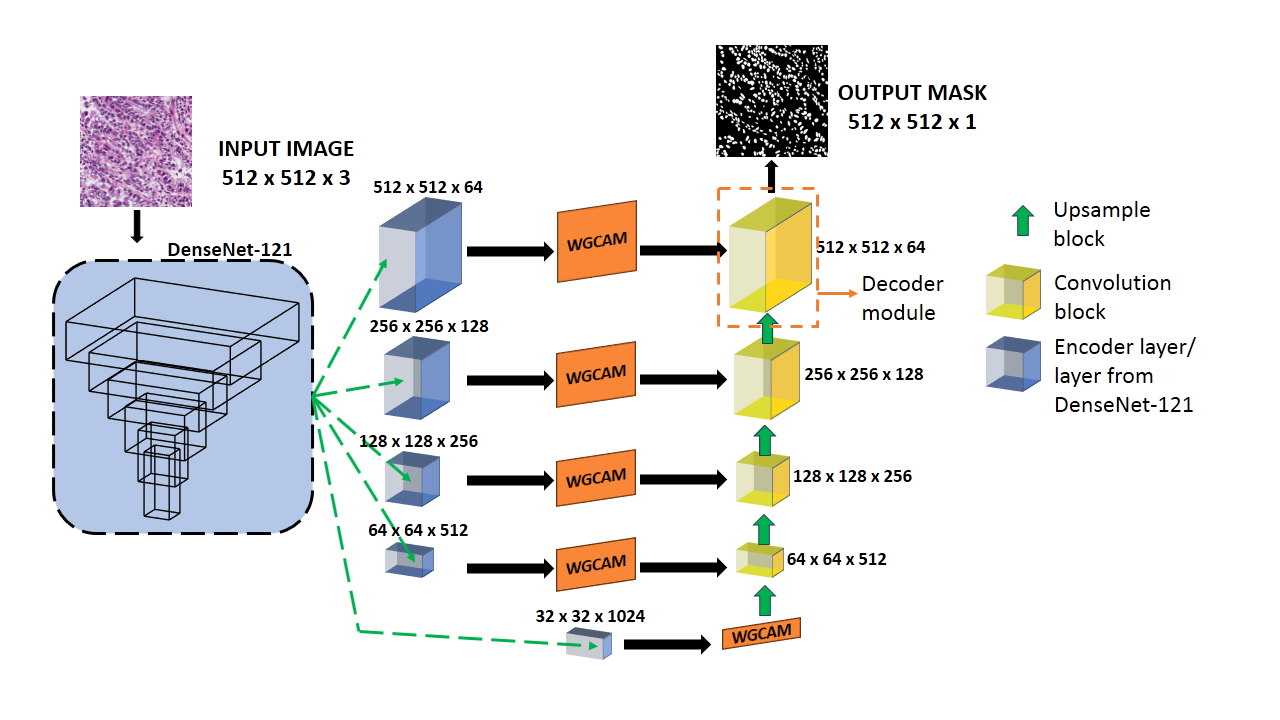}
    \caption{Our segmentation model reinforces decoder features with edge information-enhanced features using WGCAM.} 
    \label{fig:attention}
\end{figure}

The proposed segmentation model combines the structure of U-Net with the feature-rich DenseNet-121 as a backbone network to learn hierarchical features from input data. The WGCAM module captures edge information for spatial guidance, followed by a lw-GAP module for providing attention to specific channels. The decoder module comprises the upsample block and the convolution block. The upsampling block merges locally extracted features from the transposed convolution layer with upsampled features generated through Gaussian and Lanczos filters, resulting in a noise-suppressed and anti-aliased upsampled feature map. This upsampled feature map, achieved by concatenating the attention-guided features, serves as input for the convolution block. In this block, features from multiple receptive fields are extracted using convolution layers with kernels of different sizes. The pipeline of the proposed model is shown in Fig.~\ref{fig:attention}. 

\subsection{Wavelet-guided Channel Attention Module}
The WGCAM captures the edge information using wavelet transform \cite{ghazali2007feature}. A Haar wavelet is applied to the input feature, $F_{inp}$ of dimensions $H \times W \times C$ to decompose it to form $F_{wav}$ of dimensions $H/2 \times W/2 \times 4C$. $F_{wav}$ is operated with a transposed convolution to modify its dimensions to $H \times W \times 4C$. A separable convolution layer is then utilized to generate the attention weights $F'_{wav}$ of dimensions $H \times W \times C$ as shown in Eq. \ref{eq:wav}, where $f_{sc}$ and $f_{ct}$ are the separable convolution and transposed convolution, respectively. 

\begin{figure}[htb]
    \centering
    \includegraphics[width=1.0\linewidth]{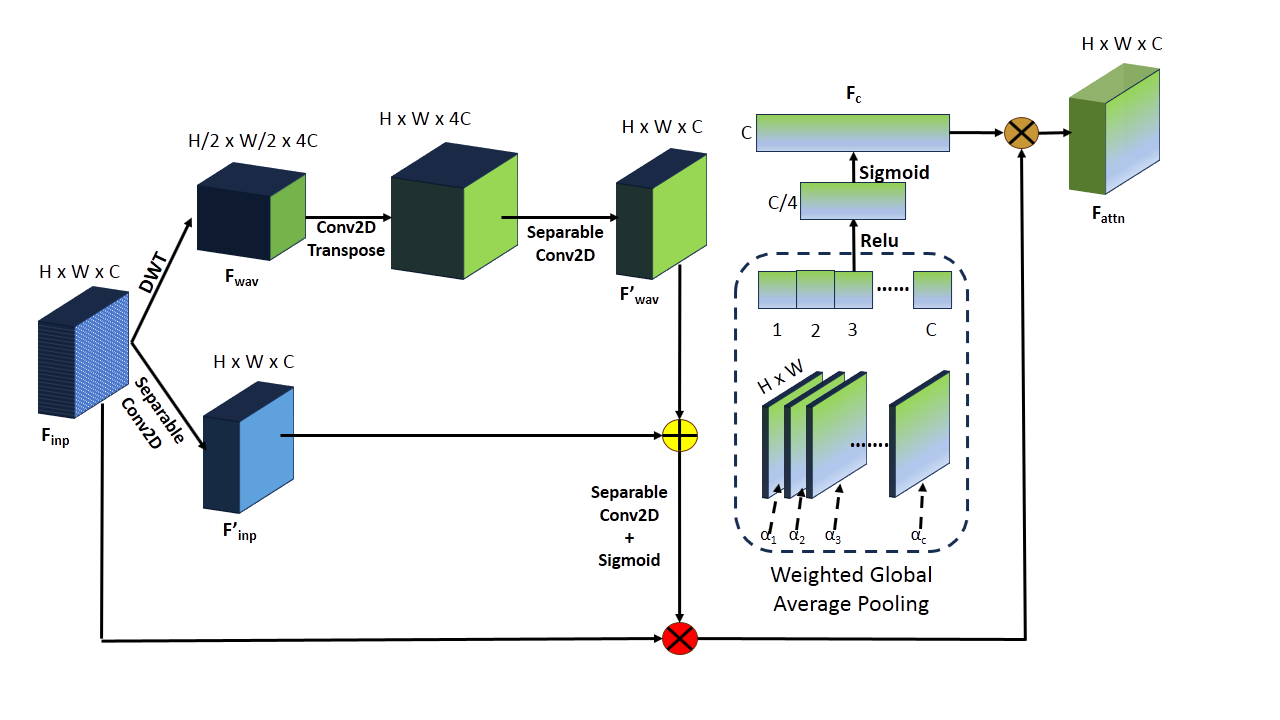}
    \caption{Wavelet-guided Channel Attention Module} 
    \label{fig:wgcam}
\end{figure}

\begin{equation}
    F'_{wav} = f_{sc}^{1 \times 1}(f_{ct}^{1 \times 1}(DWT_{haar}(F_{inp})))
    \label{eq:wav}
\end{equation}

$F'_{wav}$ is then added with $F'_{inp}$ (generated by applying separable convolution on $F_{inp}$). A separable convolution layer followed by a sigmoid activation function is utilized to generate the attention-aided feature. This attention-aided feature is then element-wise multiplied with $F_{inp}$ followed by a channel-wise multiplication with attention weights $F_c$ across the channel dimension as shown in Eq. \ref{eq:attn} where $\times$ is element-wise multiplication and $\otimes$ is the multiplication across the channel dimension. 

\begin{equation}
    F'_{wav} = F_c \otimes (Sigmoid(f_{sc}^{1 \times 1}(F'_{inp} + F'_{wav})) \times F_{inp})
    \label{eq:attn}
\end{equation}

$F_c$ is obtained by a learnable weighted Global Average Pooling (lw-GAP) where learnable weights $\alpha_{1}$, $\alpha_{2}$,... $\alpha_{C}$ are assigned to each channel. This flattened 1D tensor of dimension C then passes through two dense layers to generate $F_c$ as shown in Fig. \ref{fig:wgcam}.

\subsection{The Decoder Module}
In digital images, noise and aliasing can be problematic, especially in histopathological images, given their complex staining patterns and extremely small regions of interest. To address these challenges, we have applied the Gaussian filter \cite{lee2010nonlinear} to suppress noise and the Lanczos filter \cite{panda2022new} with a $5 \times 5$ kernel size for its anti-aliasing properties in the upsampling layers. The upsampled features are combined to create $F_{up}$, and the feature $F_{ct}$ from the transposed convolution layer offers detailed local information to compensate for any data loss during interpolation in the upsampling layer. The final feature $F_{up-ct}$ is generated by fusing features from both techniques, as illustrated in Eq. \ref{eq:comb}, where $\oplus$ denotes concatenation.

\begin{equation}
    F_{up-ct} = (f_{c}^{1 \times 1}(F_{up}) \oplus F_{ct})
    \label{eq:comb}
\end{equation}

\begin{figure}[htb]
    \centering
    \includegraphics[width=0.95\linewidth, height=5cm]{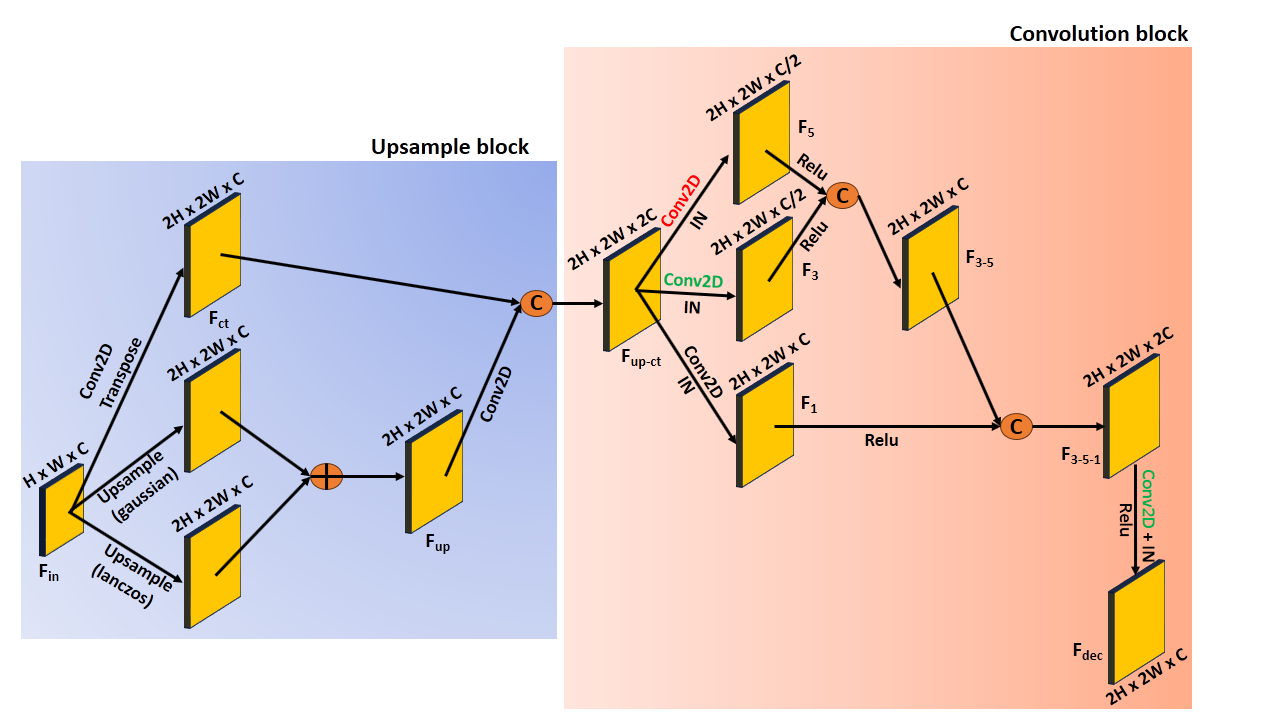}
    \caption{A block diagram of the Decoder module. The two key components are the upsample and convolution blocks} 
    \label{fig:updec}
\end{figure}

The convolution block takes the input feature map $F_{up-ct}$ and extracts information from three different receptive fields. It uses convolution layers with kernel sizes of $5 \times 5$ ($F_{5}$), $3 \times 3$ ($F_{3}$), and $1 \times 1$ ($F_{1}$). These features ($F_{5}$, $F_{3}$, and $F_{1}$) are normalized using Instance Normalization (IN) ~\cite{ulyanov2016instance} to account for variations among images in a dataset, followed by a ReLU activation. The normalized features are then concatenated and passed through convolution layers, instance normalization, and Rectified Linear Unit (ReLU) activation to produce the decoded feature map $F_{dec}$ as shown in Fig. \ref{fig:updec}. 

\section{Results}
\subsection{Experimental Setup}
We have evaluated our model on two datasets: MonuSeg~\cite{kumar2019multi} and TNBC~\cite{naylor2018segmentation}. MonuSeg contains $512\times512$ Hematoxylin and Eosin-stained tissue images, with 30 training images (22,000 annotations) and 14 test images (7000 annotations). The TNBC dataset focuses on Triple-Negative Breast Cancer tissues, with 50 images (4022 annotated cells). The original image size ($512\times512$) was used as input for both datasets. We trained using a 70-20-10 \% train-validation-test split, using a learning rate of 0.0001, the Adam optimizer, batch size of 2, and training for 100 epochs. We have used dice loss and binary cross-entropy (BCE) loss \cite{jadon2020survey} for training and evaluated using dice, Intersection over Union (IoU), precision and recall as standard metrics.

\subsection{Ablation study}
To figure out the optimal setup and parameters for our model, we have performed an extensive ablation study on the MonuSeg dataset. The experiments are listed below:

(i) U-Net with DenseNet-121 as the backbone

(ii) (i) + WGCAM with GAP 

(iii) (i) + lw-WGCAM with GAP 

(iv) The proposed model: (iii) + the decoder module

\begin{figure*}[htb]
     \centering
     \subfloat[TNBC \label{fig:seg_tnbc}]{%
      \includegraphics[width=0.48\textwidth, height=4.4cm]{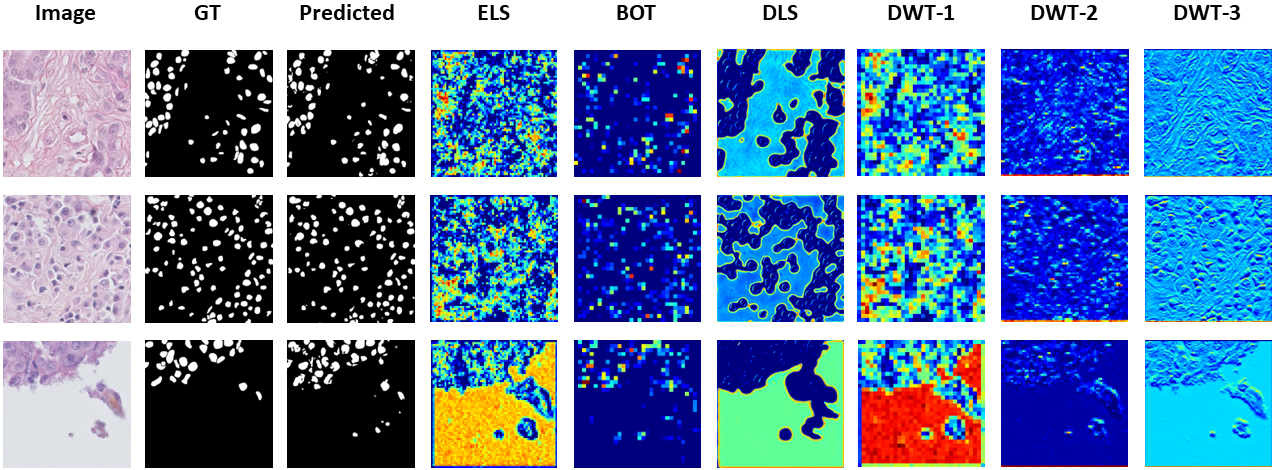}}
     \hfill
      \subfloat[Monuseg \label{fig:seg_monu}]{%
      \includegraphics[width=0.48\textwidth, height=4.4cm]{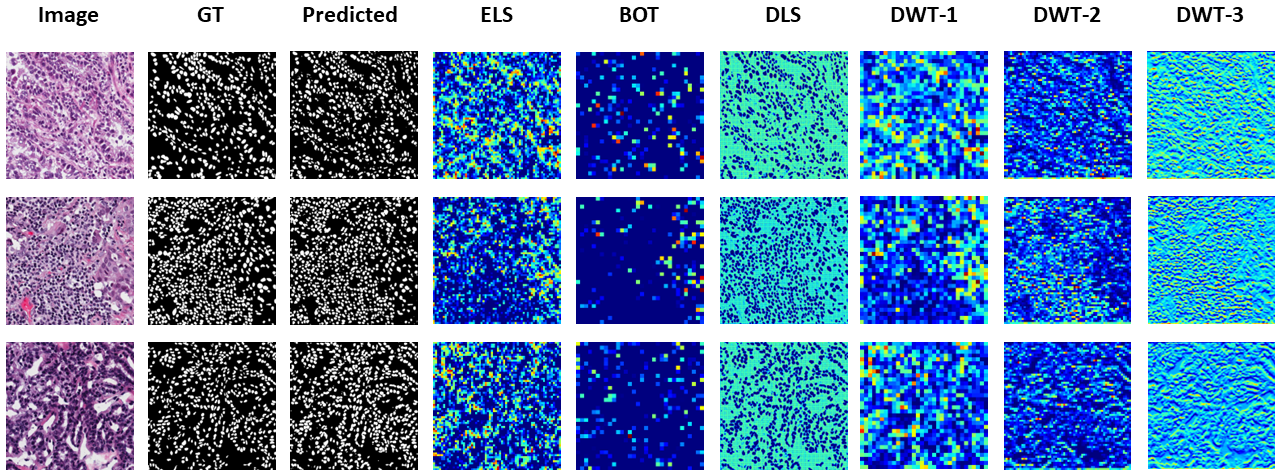}}
      \caption{Segmentation results on MonuSeg and TNBC datasets. GT represents ground truth, ELS, BOT, and DLS are the Encoder Last Layer, BOTtleneck, and Decoder Last Layer, respectively. DWT-1, DWT-2, and DWT-3 are the wavelet features of WGCAM between the second, third and fourth encoder and decoder layers, respectively.}
     \label{fig:segmask}
\end{figure*}

Table~\ref{ablation1} highlights the substantial impact of WGCAM attention along with the decoder and the upsample module on the performance enhancement of U-Net with DenseNet-121 as the backbone. The denoising and anti-aliasing capacity of the upsample layers proves instrumental in tackling the noisy and complex straining patterns of the images. Also, arbitrary shapes are easily detected with the help of edge information obtained from the wavelet features of the WGCAM. Fig. \ref{fig:segmask} shows the segmentation results and the impact of each module of the proposed model on both datasets. We compare our method with state-of-the-art (SOTA) methods on both datasets and tabulate the results in Table \ref{sota}.

\begin{table}[htb]
    \centering
    \caption{Performance of the segmentation models. All values are in \%. Bold values indicate superior performance.}
    \begin{tabular}{cccccc}        
        \textbf{Model} & \textbf{Dice} & \textbf{Precision} & \textbf{Recall}  & \textbf{IoU}\\
        \hline
        (i) & 77.39 & 72.56 & 83.17 & 63.17 \\
        (ii) & 78.77 & 73.67 & 84.89 & 65.03 \\
        (iii) & 78.89 & 75.75 & 82.62 & 65.20 \\
        (iv) & \textbf{79.46} & \textbf{76.26} & \textbf{84.91} & \textbf{66.57} \\
        \hline
    \label{ablation1}
    \end{tabular}
\end{table}

\begin{table}[htb]
    \caption{Performance comparison of the proposed model with SOTA methods. All values are in \%. Bold values indicate superior performance.}
    \begin{tabular}{|l|l|l|l|l|} 
        \hline
        \multirow{2}{*}{\textbf{Model}} & \multicolumn{2}{c|}{\textbf{Monuseg}} & \multicolumn{2}{c|}{\textbf{TNBC}} \\ \cline{2-5}
                                        & \textbf{Dice} & \textbf{IoU} & \textbf{Dice} & \textbf{IoU} \\ \hline
        U-Net\cite{ronneberger2015u}    & 74.67 & 60.89 & 68.61 & 52.92\\ \hline
        Attention U-Net\cite{oktay2018attention} & 78.67 & 66.51 & 71.43 & 54.21\\ \hline
        DIST\cite{naylor2018segmentation} & 77.31 & 63.77 & 70.51 & 56.34\\ \hline
        MMPSO-S~\cite{kanadath2023multilevel} & 72.00 & 56.00 & 65.00 & 49.00 \\ \hline
        Deep-Fuzz~\cite{das2023deep} & 79.10 & 66.10 & 77.80 & 64.20 \\ \hline
        \textbf{Proposed Method} & \textbf{79.46} & \textbf{66.57} & \textbf{81.65} & \textbf{69.18}\\ \hline
    \end{tabular}
    \label{sota}
\end{table}

\section{Conclusion}
We present a novel histopathological image segmentation model, which comprises three essential modules: WGCAM, lw-GAP and the decoder. WGCAM effectively captures edge information, improving precise cell boundary delineation. The introduction of lw-GAP enhances performance by providing channel-specific attention. Within the decoder module, the upsample block reduces noise and aliasing in the upsampled feature map by combining locally extracted features with upsampled features generated through Gaussian and Lanczos filters. This noise-reduced feature map is further strengthened by concatenating attention-guided features, significantly improving boundary delineation. The convolution block utilizes convolution layers with varying kernel sizes, adapting to diverse image variations and enhancing feature extraction. 
Future work will explore the model's performance across other medical image modalities.

  


\bibliographystyle{IEEEbib}
\bibliography{Manuscript}

\end{document}